\documentclass[11pt,a4paper]{article}

\usepackage[style=authoryear,dashed=false,backend=bibtex]{biblatex}
\bibliography{profanity}

\usepackage{times}
\usepackage{latexsym}
\usepackage{graphicx}
\usepackage{url}
\usepackage{xspace}
\usepackage{lingmacros}
\usepackage{marginnote}
\usepackage{color,soul}
\usepackage[defaultlines=4,all]{nowidow}

\usepackage[top=2.5cm,bottom=2.5cm,left=2.1cm,right=2.1cm]{geometry} 

\DeclareRobustCommand{\eg}{\textit{e.g.}\@\xspace}
\DeclareRobustCommand{\ie}{\textit{i.e.}\@\xspace}

\makeatletter
\DeclareRobustCommand{\etc}{
    \@ifnextchar{.}%
        {etc}%
        {etc.\@\xspace}%
}
\makeatother

\makeatletter
\renewcommand{\@marginparreset}{%
  \reset@font\scriptsize
  \raggedright
  \@setminipage
}
\makeatother

\DeclareRobustCommand{\ngs}{$n$-grams\@\xspace}
\DeclareRobustCommand{\hatec}{\textsc{Hate}\@\xspace}
\DeclareRobustCommand{\offnc}{\textsc{Offensive}\@\xspace}
\DeclareRobustCommand{\okc}{\textsc{Ok}\@\xspace} 

\title{\bf{Challenges in Discriminating Profanity from Hate Speech}}

\author{Shervin Malmasi\textsuperscript{1}, Marcos Zampieri\textsuperscript{2}\\
	    \textsuperscript{1}Harvard Medical School, Boston, MA 02115, USA \\
	    \textsuperscript{2}University of Wolverhampton, United Kingdom\\
	    {\tt smalmasi@bwh.harvard.edu, m.zampieri@wlv.ac.uk}\\
}

\date{}

\let \newcite \textcite
\let \cite \parencite

\begin{document}

\maketitle

\section*{Abstract}

In this study we approach the problem of distinguishing general profanity from hate speech in social media, something which has not been widely considered.
Using a new dataset annotated specifically for this task, we employ supervised classification along with a set of features that includes \ngs, skip-grams and clustering-based word representations.
We apply approaches based on single classifiers as well as more advanced ensemble classifiers and stacked generalization, achieving the best result of $80\%$ accuracy for this 3-class classification task.
Analysis of the results reveals that discriminating hate speech and profanity is not a simple task, which may require features that capture a deeper understanding of the text not always possible with surface \ngs.
The variability of gold labels in the annotated data, due to differences in the subjective adjudications of the annotators, is also an issue. Other directions for future work are discussed.
\\
\vspace{2mm}

\noindent {\bf Keywords:} hate speech, social media, bullying, Twitter, text classification, classifier ensembles

\section{Introduction}

With the rise of social media usage and user-generated content in the last decade, research into the safety and security issues in this space has grown. This research has included work on issues such as the identification of cyber-bullying \cite{xu2012learning}
and the detection of hate speech \cite{burnap2015cyber}, amongst others.

Bullying is defined as the intimidation of a particular individual while hate speech is considered the denigration of a group of people. A common thread among this body of work is the use of profane language. Both hate speech and bullying often include frequent use of profanities and research in this area has widely relied on the use of such features in building classification systems. However, the use of offensive language can occur in other contexts as well, including informal communication in non-antagonistic scenarios. Consider the following example:

\enumsentence{Holy shit, look at these ****** prices... damn!}

\noindent Most studies conducted in these areas, as we discuss in Section 2, are conducted as binary classification tasks with positive and negative classes, \eg for bullying vs non-bullying. Classifiers trained on such data often rely on the detection of profanity to distinguish between the two classes. For example, in analyzing the errors committed by their bullying classifier, \newcite{dinakar2011modeling} note that ``the lack of profanity or negativity [can] mislead the classifier" in some cases. To address such issues, we need to investigate how such classification systems perform for discriminating between the target class (\ie bullying or hate speech) and general profanity.
The distinction here is that general profanity, like in the example above, is not necessarily targeted towards an individual and may simply be used for emphasis.

The goal of such research is the development of systems to assist in filtering hateful or vitriolic content from being published.
From a practical perspective, the filtering of all items containing offensive language may not be desirable, given the prevalent use of profanity in informal conversation and the general uncensored and unrestricted nature of the Internet. Ideally, we would like to be able to separate these two classes. This separation of non-insulting profane language from targeted insults has not been addressed by current research.

Accordingly, the overarching aim of the present study is analyze the performance of standard classification algorithms for these tasks for distinguishing general profanity from targeted hate speech attacks.
We aim to do this by
(1) identifying and describing an appropriate language resource for this task;
(2) conducting classification experiments on this data using a variety of features and methods; and
(3) assessing performance and identifying potential issues for future work.

The remainder of this paper is organized as follows. In section 2 we describe some previous work. Our data is introduced in section 3, followed by an outline of our features in section 5. The experimental methodology is laid out in section 5, and our three experiments and results are described in section 6--8. A feature analysis is presented in section \ref{sec:featureanalysis} and  we conclude with a discussion in section \ref{sec:discussion}.

\section{Background}
\label{sec:bg}

The interest in detecting bullying and hate speech, particularly on social media, has been growing in recent years. This topic has attracted attention from researchers interested in linguistic and sociological features of hate speech, and from engineers interested in developing tools to deal with hate speech on social media platforms. In this section we review a number of studies and briefly discuss their findings. For a recent and more comprehensive survey on hate speech detection we recommend \newcite{schmidt2017survey}.

\newcite{xu2012learning} apply sentiment analysis to detect bullying roles in tweets. They also use Latent Dirichlet Allocation \cite{blei2003latent} to identify relevant topics in the bullying texts.
The detection of bullying is formulated as a binary classification task in their work, \ie the text is classified as an instance of bullying or not.

\newcite{dadvar2013improving}
improve the detection of bullying by utilizing user features and the author's frequency of profanity use in previous messages. While this is useful overall, it may not work in scenarios where the author's past writings are unavailable or if they regularly use profanity.

In an attempt to ``locate the hate", 
\newcite{kwok2013locate} collect two classes of tweets and use unigrams to distinguish them. \newcite{djuric2015hate} also build a binary classifier to distinguish between hate speech and ``clean" user comments on a website.
They adopt an approach based on word embeddings, achieving slightly better performance than a bag of words model.
However, they do not distinguish between profanity and hateful comments, conflating both into a single group.

\newcite{burnap2015cyber}
study cyber hate on Twitter. They annotate a collection of $2{,}000$ tweets which were annotated as being hateful or ``benign". They applied binary classification methods to this data, achieving a best F-score of $0.77$ for the task.

\newcite{nobata2016abusive} apply a computational model to analyze hate speech taking a temporal dimension into account. They analyze hate speech over the course of one year, applying different features such as syntactic features, different types of embeddings, and the standard character-based features. A logistic regression model trained on all features achieved 0.79 F-score in discriminating between abusive and non-abusive language.

Based in Critical Race Theory (CRT), \newcite{waseem2016hateful} present several criteria to identify racist and sexist expressions in English tweets. The dataset\footnote{\url{https://github.com/zeerakw/hatespeech}} created for these experiments is freely available for the research community \cite{waseem:2016:NLPandCSS}. 

Regarding available datasets, it should be pointed out that compiling and annotating datasets for the purpose of studying hate speech and profanity is a laborious and non-trivial task. There is often no consensus about the phenomena which should be annotated leading to the creation of datasets with low inter-annotator agreement. One paper that discusses this issue is the one by \newcite{ross2016measuring}. In this paper the authors measured the reliability of annotations when compiling a dataset containing Germans tweets about the European refugee crises between February and March 2016. Findings of this study conclude that ``hate speech is a vague concept that requires significantly better definitions and guidelines in order to be annotated reliably''. This is a valid observation and our paper contributes to this discussion. Throughout this paper, and most notably in Section \ref{sec:discussion}, we analyze and discuss the quality of the annotation of the dataset we are using in these experiments.

Two recent events evidence the interest of the research community in the study of abusive language and hate speech using computational methods. The first of these two events is the workshop on Text Analytics for Cybersecurity and Online Safety (TA-COS) \footnote{\url{http://www.ta-cos.org/home}} held in 2016 at the biannual conference on Language Resources and Evaluation (LREC) and the second one is the Abusive Language Workshop (AWL)\footnote{\url{https://sites.google.com/site/abusivelanguageworkshop2017/}} held in 2017 at the annual meeting of the Association for Computational Linguistics (ACL).

The vast majority of studies on hate speech and abusive language, including ours, have focused on English due to the availability of suitable annotated datasets. Some of these datasets are referred to in this section. However, a few recent studies have also been published on other languages. Examples of such studies include abusive language detection on Arabic social media \cite{mubarak2017}, a system to detect and rephrase profanity on Chinese texts \cite{su2017}, racism detection on Dutch social media \cite{tulkens2016dictionary}, and finally a dataset and annotation schema for socially unacceptable discourse in Slovene \cite{fiser2017}.

The pattern we see emerging from previous work is that they overwhelmingly rely on binary classification. In this setting, systems are trained to distinguish between hate speech (or abusive language) and texts judged to be socially acceptable. Examples of experiments modeling hate speech detection as binary classification include the aforementioned recent studies by \newcite{burnap2015cyber}, \newcite{djuric2015hate}, and \newcite{nobata2016abusive}. The use of binary classification raises the question of how such systems would perform on input that includes non-antagonistic profanity.
One of the key objectives of the current study is to assess this issue.

Another innovative aspect of our work is the use of classifier ensembles instead of single classifiers. Most previous work on this topic relied on the use of single classifiers, for example \newcite{djuric2015hate} uses a single probabilistic classifier,  \newcite{nobata2016abusive} applies a regression model. The most similar approach to ours is the one by \newcite{burnap2015cyber} which applied a meta-classifier combining outputs of three classifiers: Bayesian Logistic Regression, Random Forrest, and SVM. To the best of our knowledge, however, classifier ensembles, have not been yet tested for this task.
We chose to use classifier ensembles due their performance in similar text classification tasks. Ensembles proved to be robust methods and to obtain great performance in shared tasks such as complex word identification \cite{malmasi2016ltg} and grammatical error diagnosis \cite{xiang_et_al_2015}.

\section{Data}

For this work we utilize a new dataset by \newcite{davidson2017automated}.
The ``Hate Speech Detection" dataset is composed of $14{,}509$ English tweets. A minimum of three annotators were asked to judge each short message and categorize them into one of three classes:
(1) contains hate speech (\hatec); (2) contains offensive language but no hate speech (\offnc); or (3) no offensive content at all (\okc).

The dataset contains the text of each message along with the adjudicated label.
The distribution of the texts across the three classes is shown in Table~\ref{tab:data}. The texts are tweets and therefore limited to a maximum of $140$ characters each. 

\vspace{2mm}

\begin{table}[!ht]
\centering
\tabcolsep=0.25cm
\scalebox{1.0}{
\begin{tabular}{lr}
\hline
\textbf{Class} & \textbf{Texts} \\
\hline
\hatec	& $2{,}399$\\
\offnc	& $4{,}836$\\
\okc 	& $7{,}274$\\
\hline
Total	& $14{,}509$\\
\hline
\end{tabular}
}
\caption{The classes included in the Hate Speech Identification dataset and the number of text in each class.}
\label{tab:data}
\end{table}

\section{Features}
We use several classes of surface features, as we describe here.

\subsection{Surface \ngs}
These are our most basic features, consisting of character $n$-grams ($n=2$--$8$) and word $n$-grams ($n=1$--$3$).
All tokens are lowercased before extraction of \ngs; character \ngs are extracted across word boundaries.

\subsection{Word Skip-grams}

Similar to the above features, we also extract $1$-, $2$- and $3$-skip word bigrams.

These features were chosen to approximate longer distance dependencies between words, which would be hard to capture using bigrams alone.

\subsection{Word Representation \ngs}

We also use a set of features based on word representations.
\textit{Word representations} are mathematical objects associated with words. This representation is often, but not always, a vector where each dimension is a \textit{word feature} \cite{turian2010word}.
Various methods for inducing word representations have been proposed. These include \textit{distributional} representations, such as LSA, LSI and LDA, as well as \textit{distributed} representations, also known as \textit{word embeddings}.
Yet another type of representation is based on inducing a clustering over words, with Brown clustering \cite{brown1992class} being the most well-known method in this category. This is the approach that we take in the present study.

Recent work has demonstrated that unsupervised word representations induced from large unlabelled data can be used to improve supervised tasks, a type of semi-supervised learning.
Examples of tasks where this has been applied include: dependency parsing \cite{koo2008simple}, Named Entity Recognition (NER) \cite{miller2004name}, sentiment analysis \cite{maas2011learning} and chunking \cite{turian2010word}. 
Such an approach could also be applied to the text classification task here.
Although we only have a very limited amount of labelled data, hundreds of millions of tokens of unlabelled text are readily available to us. It may be possible to use these to improve performance on this task.

Researchers have noted a number of advantages to using word representations in supervised learning tasks.
They produce substantially more compact models compared to fully \textit{lexicalized} approaches where feature vectors have the same length as the entire vocabulary and suffer from sparsity.
They better estimate the values for words that are rare or unseen in the training data. During testing, they can handle words that do not appear in the labelled training data but are observed in the test data and unlabelled data used to induce word representations.
Finally, once induced, word representations are model-agnostic and can be shared between researchers and easily incorporated into an existing supervised learning system.

\subsubsection{Brown Clustering}
We use the Brown clustering algorithm \cite{brown1992class} to induce our word representations.
This method partitions words into a set of $c$ classes which are arranged hierarchically.
This is done through greedy agglomerative merges which optimize the likelihood of a hidden Markov model which assigns each lexical type to a single class.
Brown clusters have been successfully used in tasks such as POS tagging \cite{owoputi2013improved}  and chunking \cite{turian2010word}.
They have been successfully applied in supervised learning tasks \cite{miller2004name} and thus we also adopt their use here.

\subsubsection{Unlabelled Data}
The unlabelled data used in our experiment comes from the clusters generated by \newcite{owoputi2013improved}.
They collected $56$ million English tweets ($837$ million tokens) and used it to generate $1{,}000$ hierarchical clusters over $217$ thousand words.
These clusters are available and can be accessed via their website.\footnote{\url{http://www.cs.cmu.edu/~ark/TweetNLP/cluster_viewer.html}}

\subsubsection{Brown Cluster Feature Representation}
Brown clusters are arranged hierarchically in a binary tree. Each cluster is a node in the tree and identified by a bitstring of length $\leq 16$ that represents its unique tree path.
Some actual examples of clusters from the data we use are shown in Table~\ref{tab:clusters}. 
We observe that words in each cluster are related both syntactically and semantically. There is also semantic similarity between words in the same path (\eg the first 3 rows).

\begin{table}
\centering
\begin{tabular}{ll}
\hline
\textbf{Cluster Path} & \textbf{Top Words in Cluster} \\
\hline
\tt 111010100010	&	\small lmao lmfao lmaoo lmaooo hahahahaha lool ctfu rofl loool lmfaoo  \\
\tt 111010100011	&	\small haha hahaha hehe hahahaha hahah aha hehehe ahaha hah hahahah  \\
\tt 111010100100	&	\small yes yep yup nope yess yesss yessss ofcourse yeap likewise yepp yesh  \\
\hline
\tt 111101011000 & \small facebook fb itunes myspace skype ebay tumblr bbm flickr msn netflix \\
\hline
\tt 0011001 & \small tryna gon finna bouta trynna boutta gne fina gonn tryina fenna qone  \\
\tt 0011000 & \small gonna gunna gona gna guna gnna ganna qonna gonnna gana qunna \\
\hline
\end{tabular}
\caption{Some example of the word clusters used here. Semantic similarity between words in the same path can be observed (e.g. the first 3 rows).}
\label{tab:clusters}
\end{table}

The bitstring associated with each word can be used as a feature in discriminative models.
Additionally, previous work often also uses a $p$-length prefix of this bitstring as a feature.
When $p$ is smaller than the bitstring's length, the prefix represents an ancestor node in the binary tree and this superset includes all words below that node.
We follow the same approach here, using all prefix lengths $p~\in$~\{$2, 4, 6, \dots , 16\}$.
Using the prefix features in this way enables the use of cluster supersets as features and has been found to be effective in other tasks \cite{owoputi2013improved}.
Each word in a sentence is assigned to a Brown cluster and the features are extracted from this cluster's bitstring.

\section{Methodology}

\subsection{Preprocessing}
The texts are first preprocessed. All tokens are lowercased. Mentions, URLs and emoji are also removed.

\subsection{Classification Models}
We use a linear Support Vector Machine to perform multi-class
classification in our experiments.  In particular, we use the
LIBLINEAR\footnote{\url{https://www.csie.ntu.edu.tw/~cjlin/liblinear/}}
package \cite{LIBLINEAR} which has been shown to be efficient for text
classification problems such as this. 
For example, it has been demonstrated to be a very effective classifier for the task of Native Language Identification \cite{malmasi-dras:2015:nli,malmasi-wong-dras:2013:BEA8} which also relies on text classification methods.

We use these models for both our single model as well as ensemble experiments.
For the ensemble methods, which  we describe in the next section, a set of base classifiers will be required. In our experiments linear SVM models are used to train each of these classifiers, with each one being trained on a different feature type in order to maximize ensemble diversity.

\subsection{Ensemble Models}
\label{sec:ensemble-combiners}

In addition to single-classifier experiments, we also apply ensembles for this task. Classifier ensembles are a way of combining different classifiers or experts with the goal of improving accuracy through enhanced decision making.
They have been applied to a wide range of real-world problems and shown to achieve better results compared to single-classifier methods \cite{oza2008classifier}.
Through aggregating the outputs of multiple classifiers in some way, their outputs are generally considered to be more robust.
Ensemble methods continue to receive increasing attention from researchers and remain a focus of much machine learning research \cite{wozniak2014survey,kuncheva2014weighted}.

\begin{figure}
\centering
\includegraphics[trim=5 0 5 0,clip,width=0.65\textwidth]{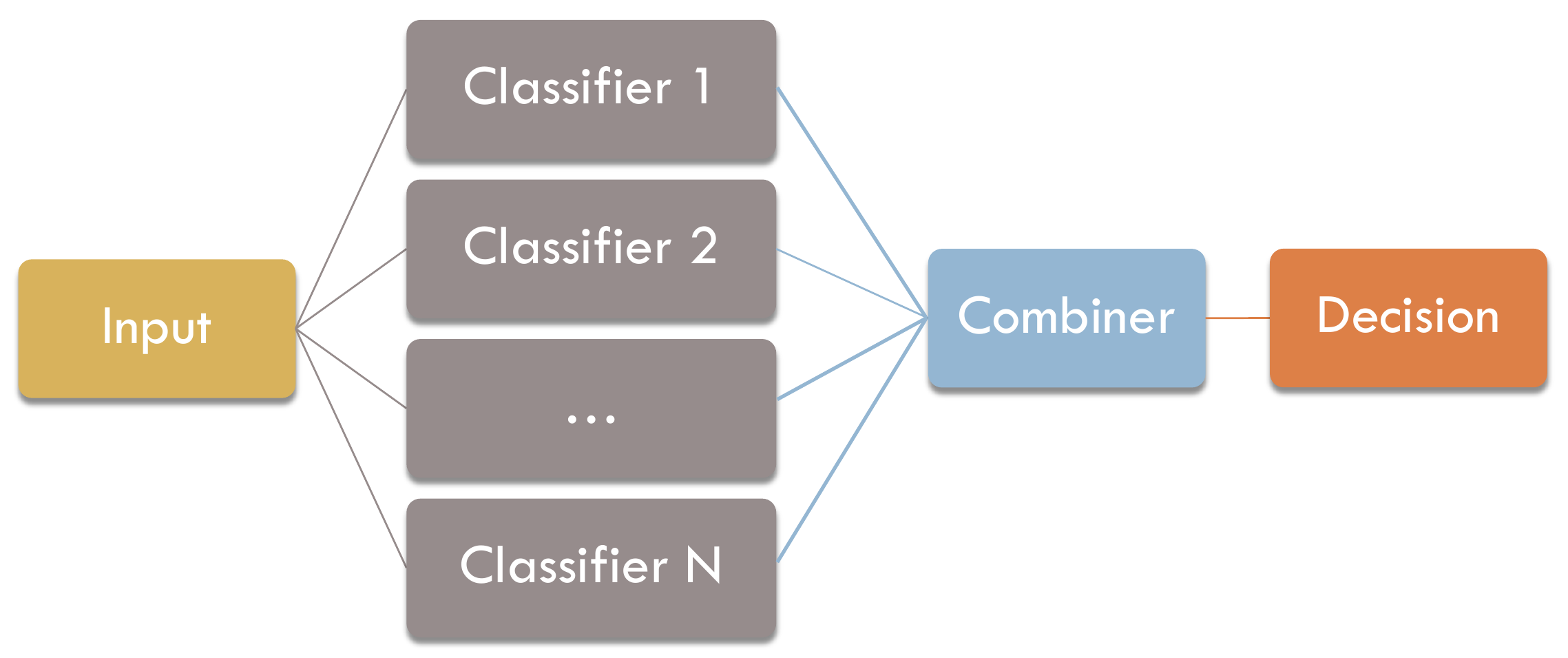}
\caption[Parallel classifier ensemble architecture]{An example of parallel classifier ensemble architecture where $N$ independent classifiers provide predictions which are then fused using an ensemble combination method. This can be considered a type of late fusion scheme. More specifically, in our experiments each classifier is a linear SVM trained on a single feature type.}
\label{fig:ensemble-architecture}
\end{figure}

Such ensemble-based systems often use a parallel architecture, as illustrated in Figure \ref{fig:ensemble-architecture}, where the classifiers are run independently and their outputs are aggregated using a fusion method.
Other, more sophisticated, ensemble methods that rely on meta-learning may employ a stacked architecture where the output from a first set of classifiers is fed into a second level meta-classifier and so on.

The first part of creating an ensemble is generating the individual classifiers.
Various methods for creating these ensemble elements have been proposed.
These involve using different algorithms, parameters or feature types; applying different preprocessing or feature scaling methods and varying (\eg  distorting or resampling) the training data.

For example, \textit{Bagging} (bootstrap aggregating) is a commonly used method for ensemble generation \cite{Breiman96baggingpredictors} that can create multiple base classifiers.
It works by creating multiple bootstrap training sets from the original training data and a separate classifier is trained from each one of these sets.
The generated classifiers are said to be diverse because each training set is created by sampling with replacement and contains a random subset of the original data.
\textit{Boosting} (\eg with the AdaBoost algorithm) is another method where the base models are created with different weight distributions over the training data with the aim of assigning higher weights to training instances that are misclassified \cite{freund1996experiments}.

Once it has been decided how the set of base classifiers will be generated,
selecting the classifier combination method is the next fundamental design question in ensemble construction.

The answer to this question depends on what output is available from the individual classifiers.
Some combination methods are designed to work with class labels, assuming that each learner outputs a single class label prediction for each data point.
Other methods are designed to work with class-based continuous output, requiring that for each instance every classifier provides a measure of confidence probability\footnote{\ie an estimate of the posterior probability for the label. For non-probabilistic classifiers the distance to the decision boundary is used for estimating the decision likelihoods.} for each class label. These outputs for each class usually sum to $1$ over all the classes.

Although a number of different fusion methods have been proposed and tested, there is no single dominant method \cite{polikar2006ensemble}.
The performance of these methods is influenced by the nature of the problem and available training data, the size of the ensemble, the base classifiers used and the diversity between their outputs.

The selection of this method is often done empirically.
Many researchers have compared and contrasted the performance of combiners on different problems, and most of these studies -- both empirical and theoretical -- do not reach a definitive conclusion \cite[p 178]{kuncheva2014combining}.

In the same spirit, we experiment with several information fusion methods which have been widely discussed in the machine learning literature.
Our selected methods are listed below.
Various other methods exist and the interested reader can refer to the exposition by \newcite{polikar2006ensemble}.

\subsubsection{Plurality voting} Each classifier votes for a single class label. The votes are tallied and the label with the highest number\footnote{This differs with a \textit{majority} voting combiner where a label must obtain over $50\%$ of the votes to win. However, the names are sometimes used interchangeably.} of votes wins. Ties are broken arbitrarily.
This voting method is very simple and does not have any parameters to tune.
An extensive analysis of this method and its theoretical underpinnings can be found in the work of \newcite[p. 112]{kuncheva2004combining}.

\subsubsection{Mean Probability Rule}
The probability estimates for each class are added together and the class label with the highest average probability is the winner.
This is equivalent to the probability sum combiner which does not require calculating the average for each class.
An important aspect of using probability outputs in this way is that a classifier's support for the true class label is taken in to account, even when it is not the predicted label (\eg it could have the second highest probability).
This method has been shown to work well on a wide range of problems and, in general, it is considered to be simple, intuitive, stable \cite[p. 155]{kuncheva2014combining} and resilient to estimation errors \cite{kittler1998combining} making it one of the most robust combiners discussed in the literature.

\vspace{2mm}

\begin{figure}[!ht]
\centering
\includegraphics[width=0.60\textwidth]{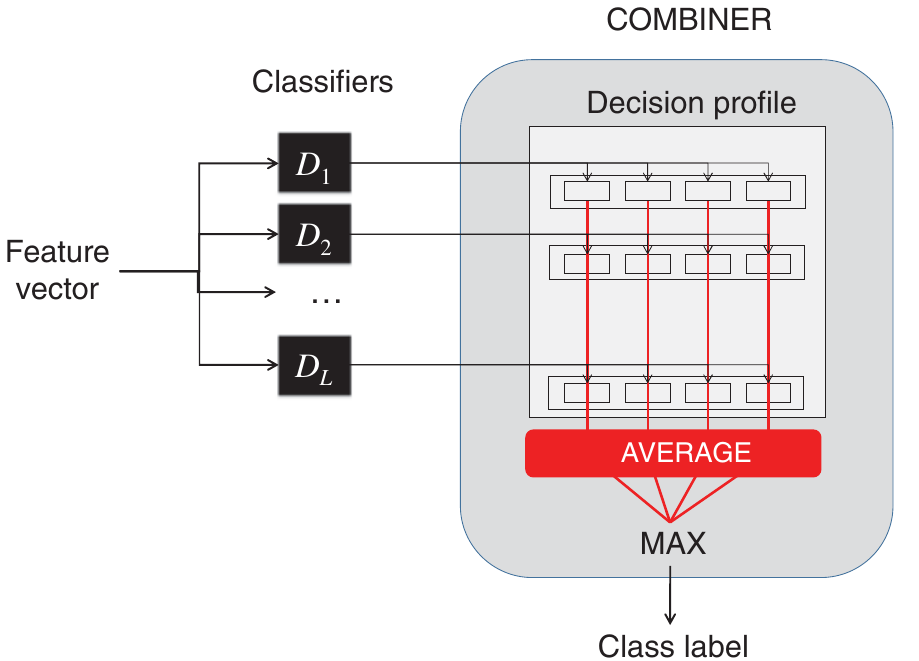}
\caption[]{An example of a mean probability combiner. The feature vector for a sample is input to $L$ classifiers, each of which output a vector of confidence probabilities for each possible label. These vectors are combined to form the decision profile for the instance which is used to calculate the average support for each label. The label with the maximum support is chosen as the prediction. This can be considered a type of late fusion. Image reproduced from \cite{kuncheva2014combining}.}
\label{fig:ensemble-average-combiner}
\end{figure}

\subsubsection{Median Probability Rule}
Given that the mean probability used in the above rule is sensitive to outliers, an alternative is to use the median as a more robust estimate of the mean \cite{kittler1998combining}.
Under this rule each class label's estimates are sorted and the median value is selected as the final score for that label. The label with the highest median value is picked as the winner.
As with the mean combiner, this method measures the central tendency of support for each label as a means of reaching a consensus decision.

\subsubsection{Borda Count}
This method works by using each classifier's confidence estimates to create a ranked list of the class labels in order of preference, with the predicted label at rank $1$.
The winning label is then selected using the Borda count\footnote{This method is generally attributed to Jean-Charles de Borda ($1733$--$1799$), but evidence suggests that it was also proposed by Ramon Llull ($1232$--$1315$).} algorithm
\cite{ho1994decision}.
The algorithm works by assigning points to labels based on their ranks.
If there are $N$ different labels, then each classifiers' preferences are assigned points as follows: the top-ranked label receives $N$ points, the second place label receives $N-1$ points, third place receives $N-2$ points and so on with the last preference receiving a single point.
These points are then tallied to select the winner with the highest score.

The most obvious advantage of this method is that it takes into account each classifier's preferences, making it possible for a label to win even if another label received the majority of the first preference votes.

\subsection{Evaluation}
We report our results as classification accuracy under $k$-fold cross-validation, with $k = 10$.
For creating our folds, we employ stratified cross-validation which aims to ensure that the proportion of classes within each partition is equal \cite{kohavi:1995}.

These results are compared against a majority baseline and an oracle.
The oracle considers the predictions by all the classifiers in Table~\ref{tab:results} and will assign the correct class label for an instance if at least one of the the classifiers produces the correct label for that data point.
This approach can help us quantify the \textit{potential} upper limit of a classification system's performance on the given data and features \cite{malmasi-tetreault-dras:2015}.

\section{Single Classifier Experiments}
\label{sec:singleclassifier}

Our first experiment aims to assess the efficacy of our features for this task.
We train a single classifier using each of our feature spaces and assess its performance under cross-validation. Additionally, we also train a single classifier by combining all of our features into single space.

\subsection{Results}

The results for our first experiment are listed in Table~\ref{tab:results}.
We observe that character \ngs perform well for this task, with $4$-grams achieving the best performance of all features.

Word unigrams also do similarly well, while performance degrades with bigrams, trigrams and skip-grams. The \ngs based on word representations also do not perform significantly better than the other features. However, these classifiers were much more efficient as they used far fewer features than word \ngs, for example.

\begin{table}[!ht]
\renewcommand{\arraystretch}{1.0}
\center
\scalebox{1.0}{
\begin{tabular}{lr}
\hline
\textbf{Feature} & \textbf{Accuracy (\%)} \\
\hline
Majority Class Baseline				& $50.1$ \\
Oracle								& $91.6$ \\
\hline
Character bigrams		& $73.6$ \\
Character trigrams		& $77.2$ \\
Character $4$-grams		& $\mathbf{78.0}$ \\
Character $5$-grams		& $77.9$ \\
Character $6$-grams		& $77.2$ \\
Character $7$-grams		& $76.5$ \\
Character $8$-grams		& $75.8$ \\
\hline
Word unigrams			& $77.5$ \\
Word bigrams			& $73.8$ \\
Word trigrams			& $67.4$ \\
\hline
$1$-skip Word bigrams			& $74.0$ \\
$2$-skip Word bigrams			& $73.8$ \\
$3$-skip Word bigrams			& $73.9$ \\
\hline
Brown cluster unigrams			& $75.0$ \\
Brown cluster bigrams			& $73.9$ \\
Brown cluster trigrams			& $68.7$ \\
\hline
All features combined 	& $77.5$ \\
\hline
\end{tabular}
}
\caption{Classification results on the dataset using various feature spaces under $10$-fold cross-validation.}
\label{tab:results}
\end{table}

The combination of all feature classes results in a very large dimensionality increase, with a total of $5.5$ million features. However, this classifier does not do significantly better than the individual ones.

\begin{figure}[!ht]
\centering
\includegraphics[trim=5 10 5 10,clip,width=0.62\textwidth]{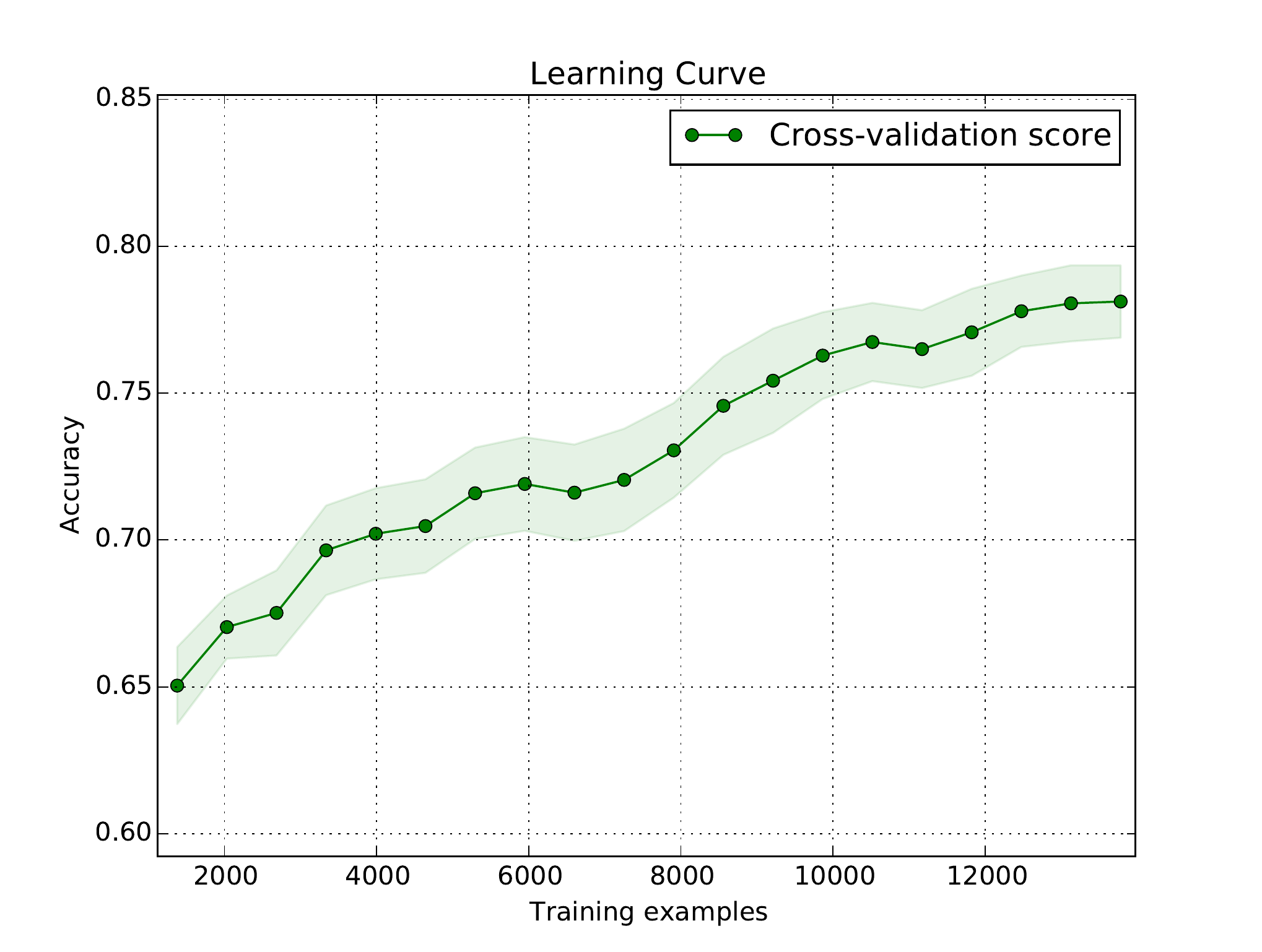}
\caption{A learning curve for a classifier trained on Character $4$-grams. The standard deviation range is also highlighted. The accuracy does not plateau with the maximal training data used.}
\label{fig:learning-curve}
\end{figure}

\noindent We also analyze the rate of learning for these features. A learning curve for a classifier trained on character trigrams and word unigrams is shown in Figure \ref{fig:learning-curve}.
We observed that accuracy increased continuously as the amount of training data increased, and the standard deviation of the results between the cross-validation folds decreased.
This suggests that more training data could provide even higher accuracy, although accuracy increases at a much slower rate after $15$k training sentences.

\section{Ensemble Classifier Experiments}
\label{sec:classifier-ensembles}

Given that a simple combination of all the features did not improve our performance on this task, we conduct another experiment where we create a classifier ensemble using our $16$ individual classifiers.

These ensemble methods were described in Section \ref{sec:ensemble-combiners}.
As we mentioned in Section \ref{sec:singleclassifier}, each of the base classifiers in our ensemble is trained on a different feature space, as this has proven to be effective.

\subsection{Results}

The results for the different ensemble fusion strategies are shown in Table~\ref{tab:results2}.
The mean probability combiner yieled the best performance, although this was still not greater than the best single classifier.
Among the voting methods, the Borda count outperformed simple plurality voting.

\vspace{2mm}

\begin{table}[!ht]
\renewcommand{\arraystretch}{1.0}
\center
\scalebox{1.0}{
\begin{tabular}{lr}
\hline
\textbf{Method} & \textbf{Accuracy (\%)} \\
\hline
Majority Class Baseline				& $50.1$ \\
Oracle								& $91.6$ \\
\hline
Character $4$-grams (single classifier)		& $\mathbf{78.0}$ \\
\hline
Plurality Voting		& $76.5$ \\
Mean Probability Rule	& $77.6$ \\
Median Probability Rule	& $76.4$ \\
Borda Count				& $77.1$ \\
\hline
\end{tabular}
}
\caption{Classification results on the dataset using various ensemble fusion methods to combine our individual classifiers.}
\label{tab:results2}
\end{table}

\section{Meta-classification Experiment}

In our final experiment, we perform another attempt at combining the different classifiers through the application of meta learning.

Meta learning is an approach to learn from what the other classification algorithms have learned in order to improve performance.
While there are various types of meta-learning, such as \textit{inductive transfer} or \textit{learning to learn} \cite{vilalta2002perspective}, some of these methods are aimed at combining multiple learners.
This particular type of meta-learning is also referred to as meta-classification.
Instead of combining the outputs of the various learners or experts in the ensemble through a rule-based approach, a different learner is trained on the outputs of the ensemble members.

It is common that certain classes in a  dataset are commonly confused or that some classifiers or feature types are more accurate for predicting certain instances or classes.
Meta-classification could be applied here to learn such relationships and build an algorithm that attempts to make a more optimal choice, not based just on consensus voting or average probabilities.

\begin{figure}[!ht]
\centering
\includegraphics[trim=0 0 0 0,clip,width=0.60\textwidth]{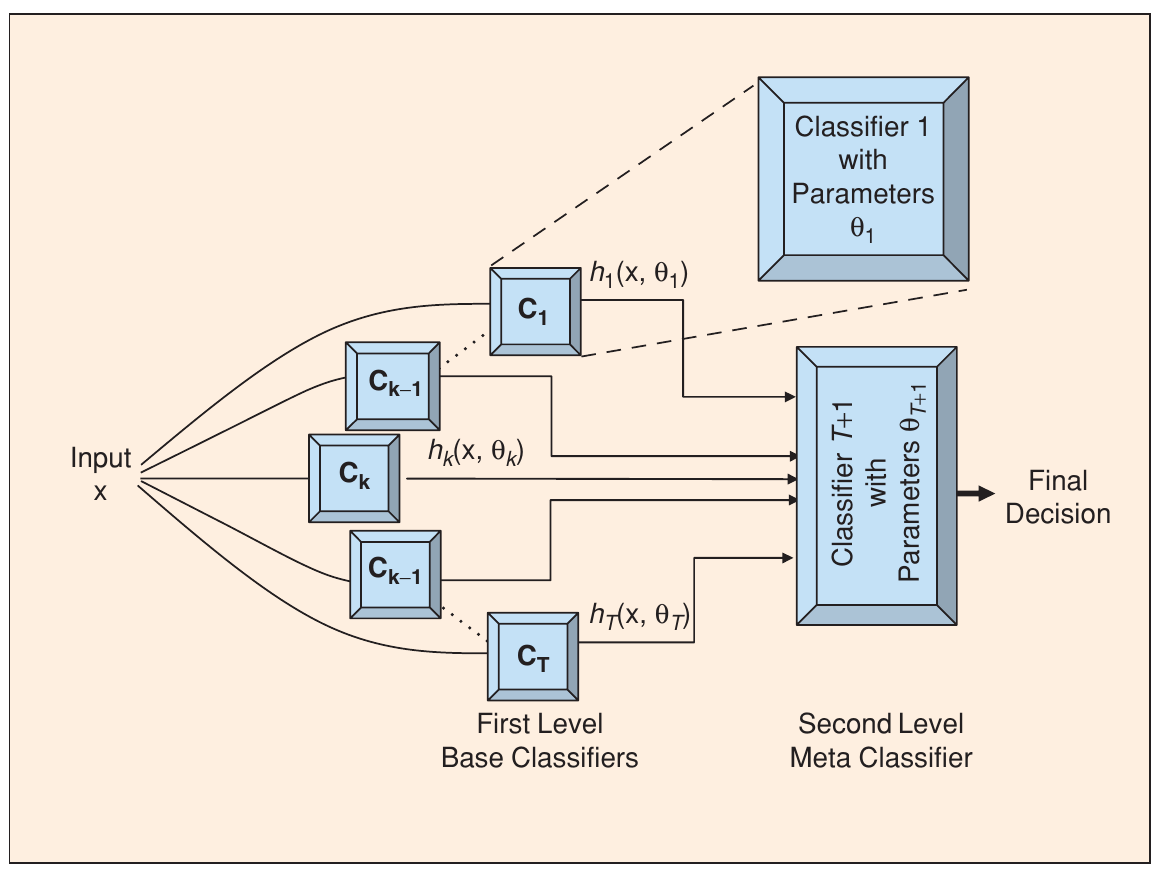}
\caption{An illustration of a meta-classifier architecture. Image reproduced from \protect\newcite{polikar2006ensemble}.}
\label{fig:metaclassifier}
\end{figure}

Stacked Generalization (also called classifier stacking) is one such meta learning method that attempts to map the outputs of a set of base classifiers for each instance to their true class labels.
To learn such a mapping, classifiers $C_1$ to $C_T$ are first trained on the the input data.
The outputs from these classifiers, either the predicted class labels or continuous output such as probabilities or confidence values for each class, are then used in conjunction with the true class labels to train a second level meta-classifier. This process is illustrated in Figure~\ref{fig:metaclassifier}.
This meta-classifier attempts to learn from the collective knowledge represented by the ensemble of local classifiers.

In our experiment, the inputs to the meta-classifier are the continuous outputs associated with each class. Each of our $16$ classifiers provides $3$ such values (one per class), for a total of $48$ features.

\noindent We experiment with two algorithms for our meta-classifier: a linear SVM just like our base classifiers and an Radial basis function (RBF) kernel SVM.
The RBF kernel SVM is more suitable for data with a smaller number of features such as here and can provide non-linear decision boundaries.

\subsection{Results}

The meta-classification results from this experiment are shown in Table~\ref{tab:results3}.

\vspace{2mm}

\begin{table}[!ht]
\renewcommand{\arraystretch}{1.0}
\center
\scalebox{1.0}{
\begin{tabular}{lr}
\hline
\textbf{Method} & \textbf{Accuracy (\%)} \\
\hline
Majority Class Baseline				& $50.1$ \\
Oracle								& $91.6$ \\
\hline
Character $4$-grams (single classifier)		& $78.0$ \\
Mean Probability Rule Ensemble		& $77.6$ \\
\hline
Linear SVM meta-classifier		& $79.0$ \\
RBF-kernel SVM meta-classifier	& $\mathbf{79.8}$ \\
\hline
\end{tabular}
}
\caption{Classification results on the dataset using two meta-classifiers to combine our individual classifiers.}
\label{tab:results3}
\end{table}

\noindent Unlike the ensemble combination methods, both of the meta-classifiers here outperform the base classifiers by a substantial margin, with the RBF kernel SVM yielding the best accuracy of $79.8\%$. This is an increase of almost $2\%$ over the best single classifier and roughly $12\%$ lower than the oracle performance.

\begin{figure}[!t]
\begin{center}
\includegraphics[width=0.52\textwidth]{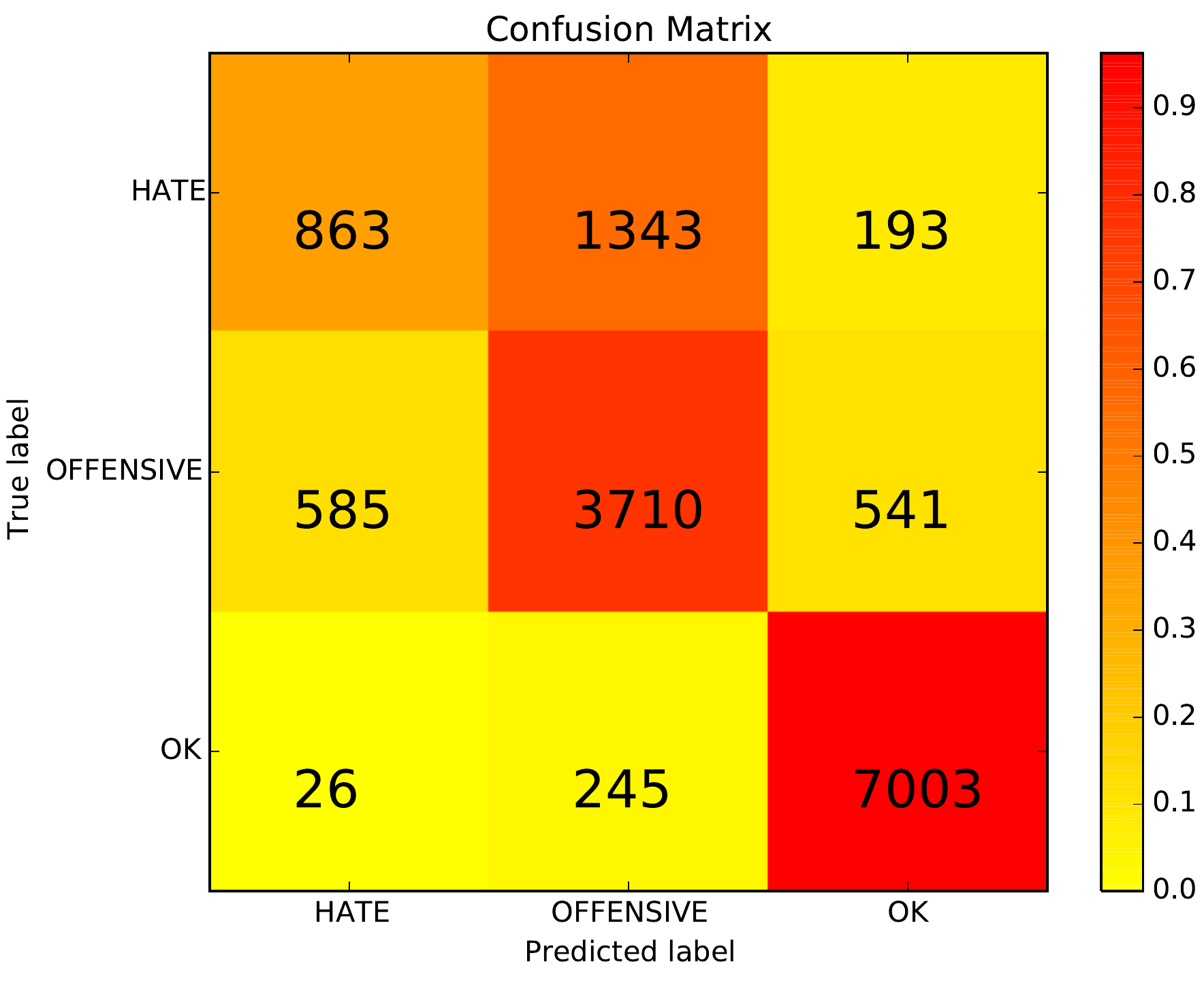}
\end{center}
\caption[]{Confusion matrix of the best result for our $3$ classes. The heatmap represents the proportion of correctly classified examples in each class (this is normalized as the data distribution is imbalanced). The raw numbers are also reported within each cell. We note that the \hatec class is the hardest to classify and is highly confused with the \offnc class.}
\label{fig:confm}
\end{figure}

A confusion matrix of the results is shown in Figure~\ref{fig:confm}. The \okc class is the most correctly classifier, with only a few instances being misclassified as \offnc.
The \hatec class, however, is highly confused with the \offnc class.

Table~\ref{tab:results4} includes the precision and recall values for each of the three classes. These results confirm that the \hatec class suffers from significantly worse performance compared to the other classes.


\begin{table}[!ht]
\centering
\tabcolsep=0.15cm
\scalebox{1.0}{
\begin{tabular}{rccc}
\hline

\textbf{Class}	& \textbf{Precision}  &  \textbf{Recall} &  \textbf{F1-score}\\
\hline
\hatec	&    0.59   &   0.36   &   0.45 \\
\offnc	&    0.70   &   0.77   &   0.73 \\
\okc	&    0.91   &   0.96   &   0.93 \\
\hline
Average	&    0.78   &   0.80   &   0.79 \\
\hline
\end{tabular}
}
\caption{The precision, recall and F1-score values for the meta-classifier results, broken down by class.}
\label{tab:results4}
\end{table}

\section{Feature Analysis}
\label{sec:featureanalysis}

In this section we look more closely at the features that help classifiers discriminate between the three classes. We take the most informative features in classification and analyze the highest ranked word unigrams and word bigrams for each class. We attempt to highlight the patterns that we consider linguistically relevant present in each class.

The most informative features were extracted using the methodology proposed in the work of \newcite{malmasi:2014:lth}. This works by ranking the features according to the weights assigned by the SVM model. In this manner, SVMs have been successfully applied in data mining and knowledge discovery in a wide range of tasks such as identifying discriminant cancer genes \cite{guyon2002gene}.

For hate speech and offensive language, we observed the prominence of profanity with coarse and obscene words being ranked as very informative for both classes. 
The main difference between the most informative features of the two classes is that posts tagged as hate speech are usually targeted at a specific ethnic or social group by using words like {\em nigger(s), jew(s), queer(s)}, and {\em faggot(s)} more often. We also observed an interesting distinction between the use of {\em nigger(s)} and {\em nigga(s)}. The first word was the second highest ranked unigram feature in hate speech and the latter was among the top twenty words in the offensive language class. This indicates that the first word and the context it is used in, is more often considered by annotators not only offensive but also a derogatory word for a target group.
This may also be related to who is using the word and its specific purpose within the discourse.

In the bigram analysis we observed that posts that were neither tagged as offensive or hate speech can be discriminated by looking at the frequency of bigrams containing grammatical words. Bigrams such as {\em and should} and {\em of those} were ranked as very informative for the {\em OK} class. In Table \ref{tab:okbigrams} we list twelve bigrams that were ranked as the most informative features for the {\em OK} class.

\vspace{5mm}

\begin{table}[!ht]
\centering
\scalebox{1.0}{
\begin{tabular}{cl}
\hline

\textbf{Rank}	& \textbf{Bigram}  \\
\hline
1	&    and should   \\
2	&    him as   \\
4	&    usually have   \\
6	&    job at   \\
7	&    marriage is   \\
9	&    are running   \\
10	&    asked if  \\
11	&    version of  \\
13	&    the rule \\
14	&    said it   \\
16	&    have another  \\
20   & 	 of those \\
\hline
\end{tabular}
}
\caption{Twelve out of top 20 highest ranked word bigrams for class {\em OK}.}
\label{tab:okbigrams}
\end{table}

\section{Discussion}
\label{sec:discussion}

A key finding of this study is the noticeable difficulty of distinguishing hate speech from profanity. These results indicate that the standard surface features applied here may not be sufficient to discriminate between profanity and hate speech with high accuracy.

Deeper linguistic processing and features may be required for improving performance in this scenario. In particular, features extracted via dependency parsing could be a promising step in this direction. In a similar vein, features extracted from a semantic parse of the text could also be used for this task, providing a representation of the meaning for each text.

We also note that the oracle performance was slightly lower than our expectations. We conducted a brief error analysis in an attempt to investigate this.
A manual review of the misclassified instances revealed that in many such cases the classifier's prediction was correct, but the assigned (gold standard) label for the text was not.
The presence of a substantial number of texts with incorrect gold labels highlights the need for better annotation. In particular, the minimum number of annotators (three for the current data) should be increased.
The provision of explicit instructions can also help as different individuals may not agree on what is offensive or not.
However, subjectivity about what is considered offensive may prove to be a challenging issue here. Additional constraints, such as the exclusion of non-native or low-proficiency speakers, could improve the quality of the judgements.
The aforementioned guidelines could be adopted for the creation of new datasets in the future, but it would also be possible to augment the existing dataset with additional judgements.

Furthermore, we also identified instances where the human judgements were correct but the classifiers could not assign the correct label. While some of these are due to insufficient training data, other cases are not as as straightforward, as demonstrated by this particular example from the data:

\enumsentence{Girls like you need to be 6 feet under}

\noindent The above text was classified as being \okc while the true label is \hatec, which we agree with.
This instance shows that not all hate speech or bullying contains profanity or colorful language; the ideas can be expressed using more subtle language.

The correct classification of this example requires a deeper understanding of the semantics of the sentence; this is likely to be difficult to achieve using surface features with supervised learning. Additional knowledge sources, including those representing idioms, may be required.

We also note that the use of hierarchical word clusters did not provide a significant performance boost.
Although they achieve similar results as other features such word \ngs, using a much small number of features.
We hypothesize that this is due to the size of the clusters used in this experiment. The $1{,}000$ clusters used here were originally induced for POS tagging of tweets.
Inspecting the clusters, we observe that while this is sufficient and effective for grouping together words of the same syntactic category, the hierarchies are not sufficiently fine-grained to distinguish the semantic groups that would be useful for our task.
For example, the words ``females" and ``dudes" are in the same cluster as several other profanities in the plural form.
Similarly, several plural insults are clustered with non-profanities such as ``iphones" and ``diamonds".
This issue can potentially be addressed by inducing a larger number of clusters, \eg $3{,}000$ or more, over similarly sized or larger data.
This is left for future work.

Finally, we did not attempt to tune the hyperparameters of our meta-classifiers; it may be possible to further improve performance by tuning these values.

It should also be noted that it is possible to approach this problem as a verification task \cite{koppel:2004:verification} instead of a multi-class or binary classification one.
In this scenario the methodology is one of novelty or outlier detection where the goal is to decide if a new observation belongs to the training distribution or not.
This can achieved using one-class classifiers such as a one-class SVM \cite{scholkopf2001estimating}.
One option is to select hate speech or bullying as the inlier training class as it may be easier to characterize this. It is also harder to define non-offensive data as there can be a much larger set of possibilities that fit into this category.

\printbibliography

\end{document}